\documentclass{article}

\usepackage{spconf,amsmath,graphicx,caption,float,placeins,endnotes, multicol,color,lipsum}

\title{EduQA: Educational Domain Question Answering System using Conceptual Network Mapping} 

\name{*A. Agarwal\textsuperscript{$\dagger$}, *N. Sachdeva\textsuperscript{$\dagger$}, *R. K. Yadav\textsuperscript{$\dagger$}, *V. Udandarao\textsuperscript{$\dagger$}, *V. Mittal\textsuperscript{$\dagger$}, A. Gupta\textsuperscript{$\dagger$}, A. Mathur\textsuperscript{$\wedge$}\thanks{*equal contribution. \newline   \{abhishek16126, nikhil16061, raj16076, vishaal16119, vrinda16279, anubha\}@iiitd.ac.in,~abhinav.mathur@millionsparks.org}}

\address{\textsuperscript{$\dagger$}SBILab, Department of ECE, IIIT-Delhi, India, \textsuperscript{$\wedge$}Million Sparks Foundation}
\begin{document}

\maketitle

\begin{abstract}
Most of the existing question answering models can be largely compiled into two categories: i) open domain question answering models that answer generic questions and use large-scale knowledge base along with the targeted web-corpus retrieval and ii) closed domain question answering models that address focused questioning area and use complex deep learning models. Both the above models derive answers through textual comprehension methods. Due to their inability to capture the pedagogical meaning of textual content, these models are not appropriately suited to the educational field for pedagogy. In this paper, we propose an on-the-fly conceptual network model that incorporates educational semantics. The proposed model preserves correlations between conceptual entities by applying intelligent indexing algorithms on the concept network so as to improve answer generation. This model can be utilized for building interactive conversational agents for aiding classroom learning. 


\end{abstract}
\begin{figure*}
\vspace{-0.5em}
    \centering
    \includegraphics[scale = 0.46]{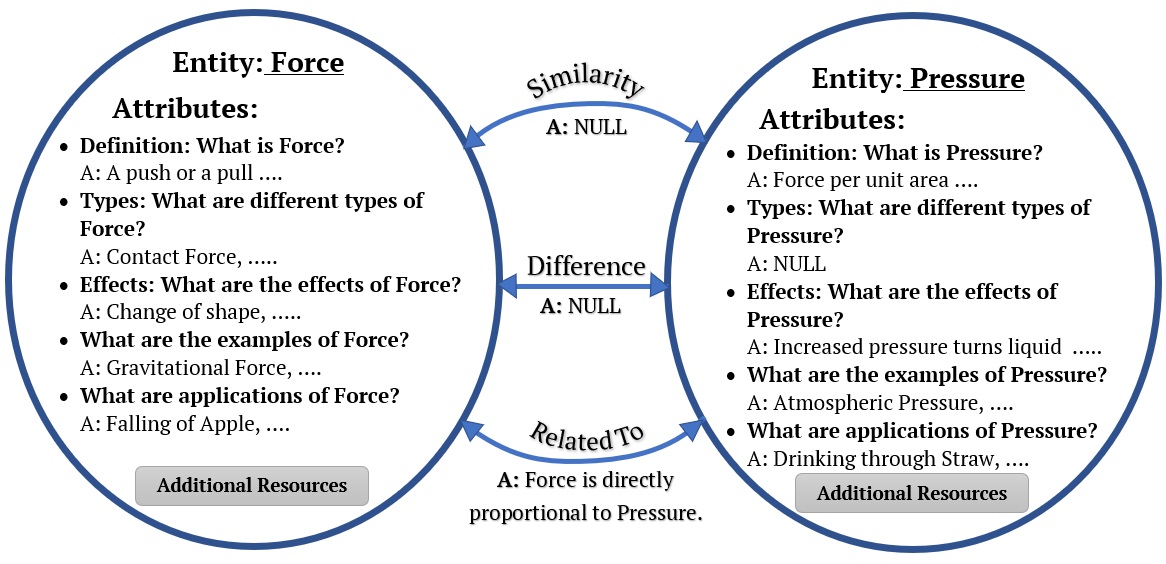}
    \vspace{-0.5em}
    \caption{\small Structure of two entities and their relationship with an example of: Force and Pressure. Here, \textbf{`A'} represents \textbf{Answer} and \textbf{`NULL'} represents that the answer does not exist.}
    \label{fig1}
    \vspace{-1.5em}
\end{figure*}
\vspace{-1em}
\section{Introduction}
\label{sec:intro}
Domain based Question Answering augments the fields of natural language processing and information retrieval, where the idea is to build a system that can autonomously answer questions from within a given field. Current question answering models are largely one of the two types, either \textit{open domain question answering models} (ODQA) answering generic questions by employing large-scale knowledge base along with the targeted web-corpus  retrieval  techniques  or \textit{closed domain question answering models} (CDQA) that use complex deep learning models  such as bidirectional attention flow using LSTMs (long short term memory recurrent neural network) to answer questions in a specific domain.

One of the popular ODQA model is such as Question Answering via Semantic Enrichment (DrQA) \cite{chen2017reading}. The problem with ODQA system such as DrQA is that even though the answer might be correct, it may not be appropriate for all types of audience because of the varying understanding level of people for different contexts. For instance, the definition of light on Wikipedia is ``Light is electromagnetic radiation within a certain portion of the electromagnetic spectrum", but for a student in $8^{th}$ grade, the definition may simply be ``light is a form of energy", as he/she would not be familiar with the concept of electromagnetic radiations. Thus, these open domain models are not suitable in the educational context. Recently, other ODQA models have been developed such as those using NLU (Natural Language Understanding) techniques to query databases \cite{braun2017evaluating}, models that retrieve answers directly from Wikipedia \cite{abbes2018towards} over multiple languages, a model that uses "Omnibase" as a knowledge base to retrieve answers \cite{katz2002omnibase}, and a wide range of models that apply the principles of IR (Information Retrieval) to provide answers \cite{Diefenbach2018}.

Some of the popular CDQA systems are Bi-directional Attention Flow (BiDAF) \cite{seo2016bidirectional} and Question Answering via Semantic Enrichment (QuASE) \cite{sun2015open}. BiDAF is a context-based question answering model that uses textual comprehension by incorporating attention mechanisms based on the top of an LSTM network and uses surface-based textual understanding techniques to derive the answers. However, it does not take into account the underlying meaning of the text. On the other hand, QuASE tries to capture the semantic similarities between the words in a question by performing a text-cum-vocabulary based semantic analysis.
A few other recent CDQA models that have been developed include a model that present answers based on pure textual contexts by document retrieval \cite{min2018efficient} and an E-Commerce restricted-domain model that uses transfer learning to efficiently retrieve answers \cite{yu2018modelling}.

However, all the existing question answering models are limited in the following ways: i) they derive their answers by rote textual learning; ii) they are unable to capture semantic correlations within the question text; iii) they are supportive of only factoid-based questions; and iv) they are not particularly useful for answering queries from students that are related to the coursework.

In addition to the above, some models exist that return Question-Specific Concept Maps \cite{atapattu2015educational}. These models focus on returning knowledge organization, specifically concept maps as answers to the question asked. They build a concept map from the resources which includes concept recognition and connection. When a question is asked, the framework is able to identify all the involved concepts but does not provide an answer. Although this model is helpful in understanding questions and its related concepts, it cannot be used as a standalone solution to an education domain question answering system in a classroom setting. 

In any educational system, gaining information is a key part of learning. Educational domain Question-Answering systems aim to provide the required explanation to the user and aid them in their learning tasks. These systems act like virtual teachers to all kinds of users. A few attempts have been made at developing such systems for the educational domain. QAPD is an ontology-based model that answers questions specifically from the physics domain by applying IR (information retrieval) on large scale ontologies \cite{abdi2018qapd}. However, QAPD answers questions for a general user and cannot be extended to the educational domain because of the different levels of understanding of students across different grades. For example, QAPD will output the same answer for the question: ``Explain gravitational force" for a 6th grade student as well as a 9th grade student although these students will have varying levels of knowledge on the same topic.

Owing to the shortcomings of the above QA models, we propose a basic framework that aims to solve the less explored problem of educational context-aware question answering. We would like to extract meaningful answers that are able to capture the pedagogical entity correlations of both the question and the answer. Such pedagogical context-aware answers can be formulated using a dynamic on-the-fly learning structure that can encode context information into an \textit{evolving network} of related entities. This structure is referred to as the \textit{Concept Network}. For example, say a teacher is teaching the $8^{\text{th}}$ grade chapter on ``Force and Pressure”. The concept network is built from the course material that was taught as part of the lessons taken. If the same chapter is taught at a later stage with some additional content, the concept network should be able to dynamically evolve to add new entities along with correlations to the pre-existing entities. Consequently, this concept network can be used to answer questions from students with a certain degree of confidence. These answers will also be contextually and semantically correct with respect to the current class $8^{\text{th}}$ chapter in focus. Thus, Our proposed model would facilitate learning in any field by enabling students to learn better by providing answers to questions with the help of a dynamic concept network that captures the descriptions and relationships among the various entities of the educational context.

\section{EduQA: Proposed Education Domain Question-Answering System}
\label{sec:majhead}
\begin{figure*}[!ht]
    \centering
    \includegraphics[scale = 0.4]{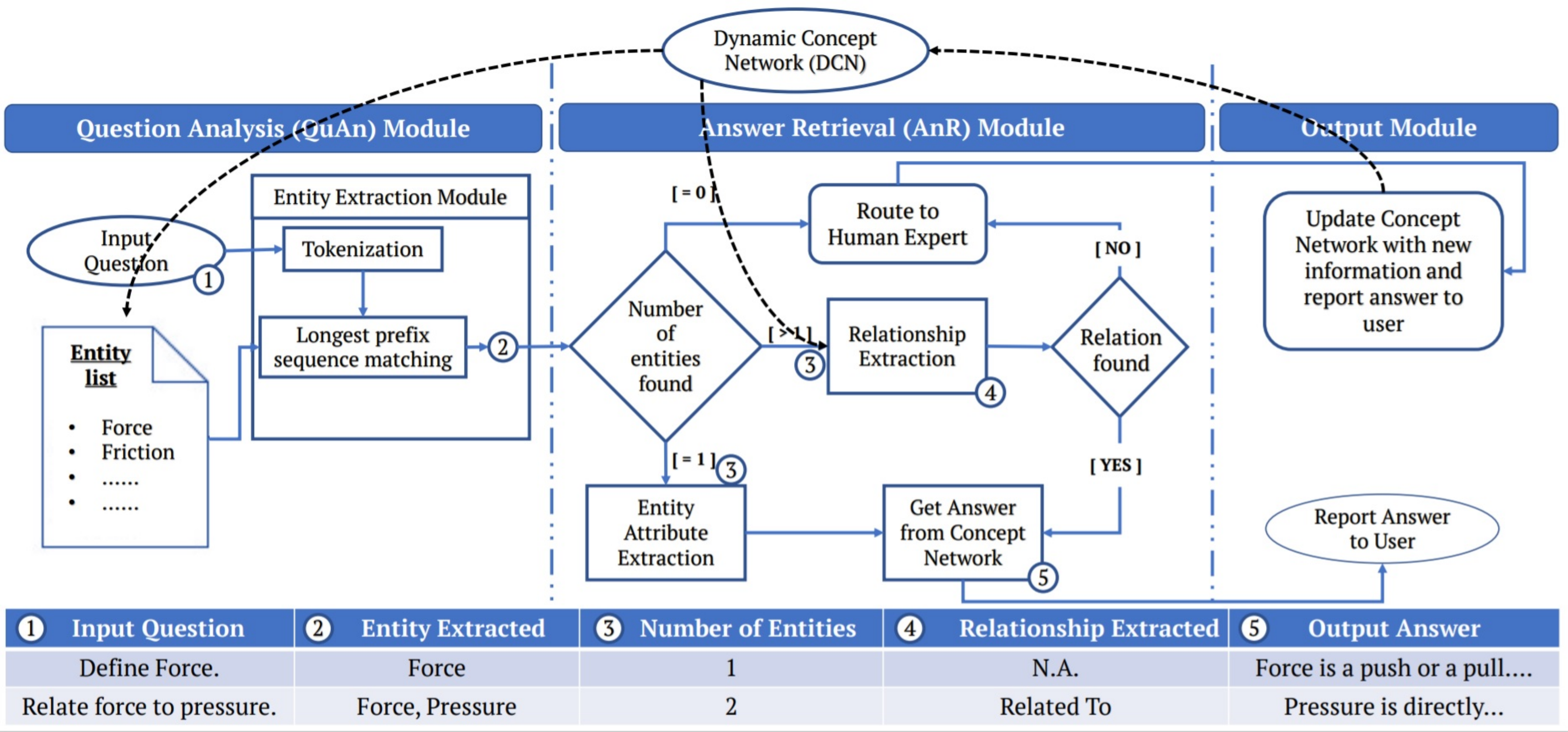}
    \caption{EduQA: Proposed Architecture for Educational Domain Question Answering}
    \label{fig:2}
\end{figure*}

The proposed architecture "EduQA" is shown in the Figure \ref{fig:2}. It consists of three modules. The first module is the Dynamic Concept Network (DCN) that consists of entities and their relationships. The second module is the Question Analysis (QuAn) that filters useful information for the answer extraction. The last module is the Answer Retrieval (AnR) that provides answer utilizing the information from the first two modules.   
\vspace{-0.5em}
\begin{figure*}
\vspace{-1em}
    \centering
    \includegraphics[scale = 0.65]{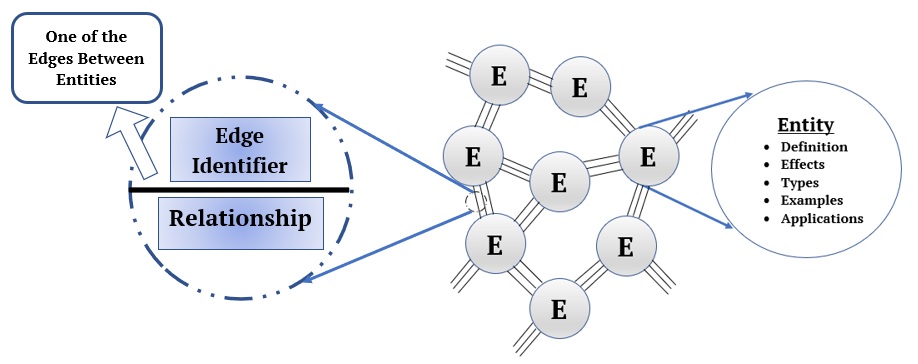}
    \caption{Generic structure of DCN identifying the structure of edges and entities.}
    \label{fig3}
    \vspace{-0.5em}
\end{figure*}
\vspace{-0.5em}
\subsection{Dynamic Concept Network (DCN) Module}
\vspace{-0.5em}
\label{ssec:subhead}
The dynamic concept network module has three components i) Entity Nodes, ii) Edges, and iii) On-the-fly learning based update. The first two components are shown in Fig.\ref{fig3}.

\textit{\textbf{Entity Nodes}}: This constitutes all the concepts included within the context. The concepts are basically the major topics or headings in the context. Each entity in the concept network is assigned to one of the topics. Further, there are a fixed number of attributes associated with every entity. These attributes store detailed and directed information about the entity in the form of a tuple. Each attribute has an identifier and a tuple. The tuple is a pair of a questions followed by its answer. The type and number of these attributes are immutable, though values for some attributes associated with some entities can be NULL. These attributes are provided in Table-1.

\textit{\textbf{Edges}}: They store the relationship between different concepts. Not all concepts are needed to be connected. The edge consists of an edge identifier and relationship. The type and number of edges between two vertices are fixed, again relationship for an edge associated with specific two vertices can be NULL. The description of edges is shown in Table-2.

\textit{\textbf{On-the-fly learning}}: The proposed “On-The-Fly” learning dynamically updates the concept network with the help of a human expert as new information is received and processed and hence, enables the model to stay updated with the advancements in the specific domain. For example, if we have a concept network of machine learning and a new machine learning model is proposed after some time, this should be reflected in the concept network. To handle this, we consider the addition of new entities and new relationships between these entities. This is done by the
\textit{addition of a new entity} as and when we come across a question that does not have any matching entity in the list of entities. In such a case, the question is forwarded to the expert and based on his input, a new entity is added to the DCN, if required. Correspondingly, \textit{a new relationship is added} When a question contains multiple entities and there is no edge (relationship) between those entities in the concept network. The question is again sent to the expert who will choose whether or not there is a need to add this relationship. This enables our model to stay updated as it is very important in the field of education.\newline
Figure \ref{fig1} shows a 2 entity subset of a concept network,built for the NCERT\footnote{National Council of Educational Research and Training (NCERT),India }grade 8 Science textbook chapter; "Force and Pressure.
\vspace{-0.9em}
\begin{figure}[H]
    \centering
    \includegraphics[scale=0.75]{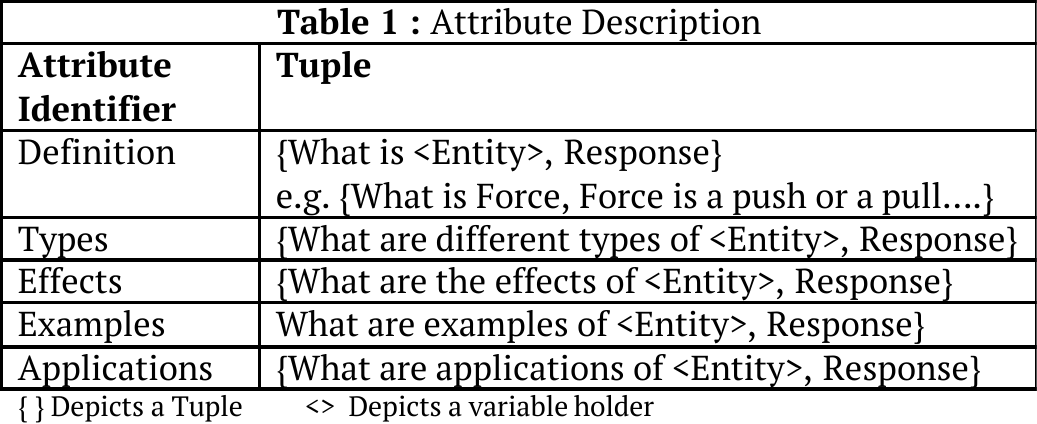}
    \label{fig:my_label}
\end{figure}
\vspace{-1em}

\begin{figure}[H]
    \vspace{-0.5em}
    \centering
    \includegraphics[scale = 0.47]{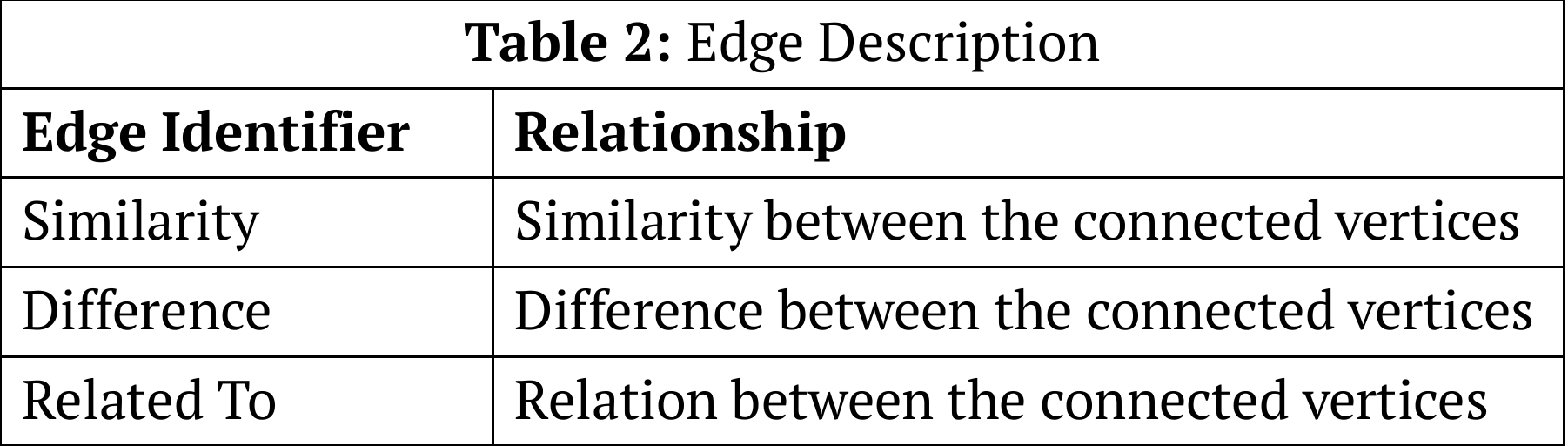}
    \label{fig:my_label}
    \vspace{-0.5em}
\end{figure}
\vspace{-2em}
\subsection{Question Analysis (QuAn) Module}
\label{ssec:subhead}
This module processes the question based on the information stored in the concept map. The details are provided below.

\textbf{\textit{Entity Extraction}: }
First, a user query is analyzed to extract the predefined concept entities from the concept network. Next, the user question is tokenized to obtain  individual words. All possible sequences of the question tokens are then constructed and fed into the "Longest Prefix Sequence Matching" component to search for the entities in the question. The final entities extracted are forwarded further in the pipeline.

The ``Longest Prefix Sequence Matching" module works in the following way. We assume that an initial dictionary that contains all possible entities as keys is available. We iterate over the entire entity set and simultaneously index each combination of the input tokens over the dictionary. If a sequence match is found between the token(s) and the entity, the corresponding entity is flagged as part of the input query. Consequently, we obtain a list of entities that are present in the question.
For instance, given the question: "What is non contact force?", we first retrieve the predefined set of concept entities from the concept network. We then proceed to tokenize the question: \{``What", ``is", ``non", ``contact", ``force"\}. Say, the set \{“non contact force”, ``contact force”, ``force”\} is a subset of the predefined concept entitites. When our entity extraction algorithm compares the token set \{``non", ``contact", ``force"\} from the question with the entity set \{``non contact force”, ``contact force”, ``force”\}, the entity ``non contact force" is returned since that is the longest prefix sequence match that can be formed using our question tokens. This entity is then sent further along the pipeline.
\subsection{Answer Retrieval (AnR) Module}
This module consists of attribute recognition and relationship extraction that finally delivers the answer.
\textbf{\textit{Attribute Recognition:}}
Once entities present in the user question are identified, questions stored in the attribute-tuples of the identified entity node are extracted. A similarity measure of the given question with each of the extracted questions \cite{song2007question} is computed as follows: 
\vspace{-0.5em}
\[
    Sim_{overall} = ({1-\delta})Sim_{statistic} + \delta{Sim_{semantic}},
\]
Statistic similarity is calculated based on
dynamically formed vectors while semantic similarity
is calculated by utilizing word similarity based on
WordNet \cite{miller1995wordnet}. Overall similarity is a combination of
statistic similarity and semantic similarity.  $\delta$ is a constant between 0 and 1. It is used to decide the contribution of the semantic similarity component in the overall similarity. Using this similarity measure, we estimate the best matched attribute question and thus, retrieve its corresponding answer for the input question.  

\textit{\textbf{Relationship Extraction:}}
Relationship extraction is done in the case when we have more than one entity in a question. This suggests that the question is asking for a relationship between those entities. In this case, we look at the edges existing between the entities and try to find the edge that has the maximum similarity with the question by comparing each edge semantically. Whichever edge matches the most, the answer stored in it will be reported.
\vspace{-1em}
\section{Results}
\vspace{-0.5em}
We compare our proposed model with the existing BiDAF and Omnibase START models. We have tested our model on a set of 50 questions taken from two NCERT\textsuperscript{$1$} class 8 chapters: "Light" and "Force and Pressure". The set of questions consist of 30 definition based questions, 5 similarity based questions, 5 difference based questions and 10 relationship questions. Out of these, the model was able to correctly answer 80\% of the definition based questions and 65\% of the other type of questions. Table-3 presents a subset of 7 questions with the corresponding answer remark on the basis of the answer given by each model. Model-1 represents our QA-system and the other systems represent existing QA-systems.
\vspace{-0.5em}
\begin{figure}[H]
\vspace{-0.5em}
    \centering
    \includegraphics[scale=0.26]{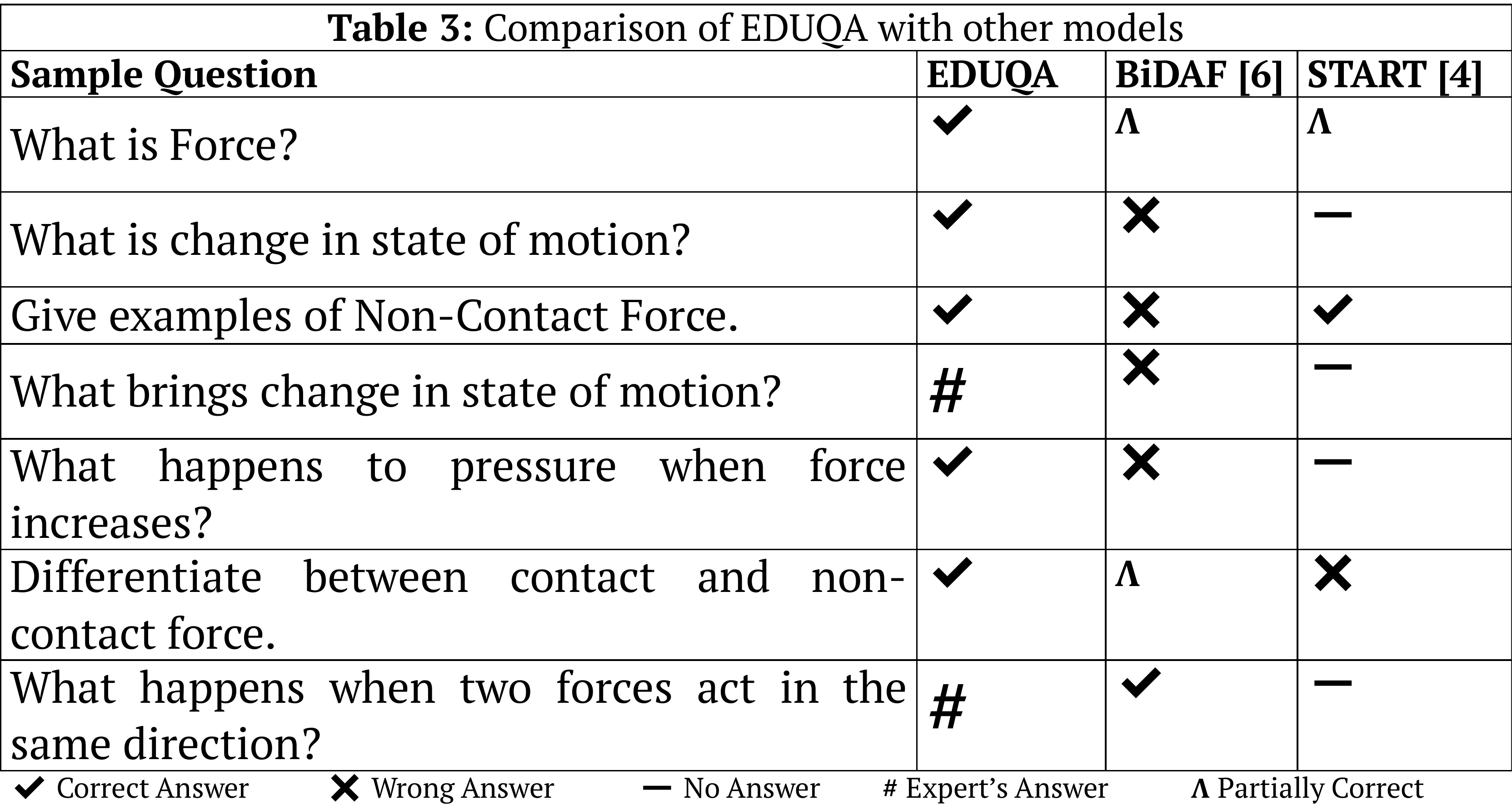}
    \label{fig10}
    \vspace{-1.5em}
\end{figure}
\section{Conclusion \& Future Scope}
\vspace{-0.5em}
In this paper, we explored the field of Question Answering with respect to an educational context. We analyzed existing frameworks and saw how they had shortcomings while answering questions related to educational domain. We proposed a framework based on a dynamic self-evolving Concept Network, built specifically for a given subject, chapter, lecture etc. In the future, we would like to: i) use better strategies for retrieval of answers from the Concept Network, ii) explore methods to best incorporate complex courses like mathematics into the concept network, iii) automate the construction of the concept network, iv) minimize the requirement of a human expert to handle unanswerable (indirect) questions, and v) include support for complex reasoning based questions.


\bibliography{bibliography}
\bibliographystyle{ieeetr}

\end{document}